\pdfoutput=1
\documentclass[10pt, a4paper]{article}

\usepackage{lrec-coling2024} % this is the new style

\usepackage{booktabs}
\usepackage{multirow}

\title{PolQA: Polish Question Answering Dataset}

\name{Piotr Rybak\textsuperscript{\textnormal{1}}, Piotr Przybyła\textsuperscript{\textnormal{1,2}}, Maciej Ogrodniczuk\textsuperscript{\textnormal{1}}
}

\address{
\textsuperscript{\textnormal{1}} Institute of Computer Science, Polish Academy of Sciences\\
ul. Jana Kazimierza 5, 01-248 Warsaw, Poland\\
\textsuperscript{\textnormal{2}} TALN Group, Universitat Pompeu Fabra\\
c. Tànger 122-140, 08018 Barcelona, Spain\\
\texttt{\{firstname.lastname\}@ipipan.waw.pl}
}

\abstract{
Recently proposed systems for open-domain question answering (OpenQA) require large amounts of training data to achieve state-of-the-art performance. However, data annotation is known to be time-consuming and therefore expensive to acquire. As a result, the appropriate datasets are available only for a handful of languages (mainly English and Chinese). In this work, we introduce and publicly release PolQA, the first Polish dataset for OpenQA. It consists of 7,000 questions, 87,525 manually labeled evidence passages, and a corpus of over 7,097,322 candidate passages. Each question is classified according to its formulation, type, as well as entity type of the answer. This resource allows us to evaluate the impact of different annotation choices on the performance of the QA system and propose an efficient annotation strategy that increases the passage retrieval accuracy@10 by 10.55 p.p. while reducing the annotation cost by 82\%.
\\ \newline 
}
\begin{document}

\maketitleabstract

\section{Introduction}
\label{sec:introduction}
% TODO: rewrite to focus more on annotation strategies
The goal of open-domain question answering (OpenQA) is to provide an answer to a question asked in a natural language. Typically, an OpenQA system consists of three components. The \emph{knowledge source} is used as a source of passages that might contain the answer. The \emph{retriever} is responsible for searching for relevant passages from the knowledge source. Finally, the \emph{reader} extracts (or generates) the answer based on the given question and retrieved passages.

Recently, neural retrieval systems (e.g. Dense Passage Retrieval, \citealp[DPR]{karpukhin-etal-2020-dense}) surpassed traditionally used lexical methods (e.g. BM-25, \citealp{Robertson2009ThePR}) by fine-tuning pre-trained language models on a large number of (question, passage) pairs. They achieve state-of-the-art results but at the cost of the necessity to annotate training sets and poor generalizability to other languages or even domains \citep{thakur2021beir}.

The third component of OpenQA systems (\emph{reader}) also requires annotated dataset for training. Besides (question, passage) pairs, it needs the final answer to the question. Reader takes the form of either a short span of the passage (extractive QA) or free text (generative QA).

In this paper, we introduce and release PolQA, the first Polish OpenQA dataset. It consists of 7,000 general knowledge questions obtained from TV quiz shows, online quizzes, and similar sources. Each question is accompanied by up to 15 evidence passages (87,525 in total) and up to five answer variants (8,713 in total). We also release a corpus of 7,097,322 candidate passages based on parsed Polish Wikipedia. Additionally, each question is described by its formulation (how the question is asked), type (what kind of information is sought), and the entity type of the answer (e.g. country, person).

The wide availability of pretrained language models means that a decently-performing system may be built with little to no training data, but adding some annotation will result in much better results. Thus, a system designer needs to predict, adding what (and how much) manually annotated data will be most cost-effective. During the course of creating PolQA, we analyzed the impact of several annotation strategies on the QA system performance and annotation cost. This allows us to answer the following research questions:
\begin{itemize}
    \item \textbf{RQ1}: What is the benefit of high-quality and unbiased training data? \vspace{-1pt}
    \item \textbf{RQ2}: Does the performance improve with more training examples? \vspace{-1pt}
    \item \textbf{RQ3}: What is the impact of manually annotated hard negative passages on different components of the OpenQA system? \vspace{-1pt}
    \item \textbf{RQ4}: What is the cost of obtaining annotations in terms of human effort? \vspace{-1pt}
    \item \textbf{RQ5}: What annotation strategy can be recommended for future OpenQA annotation efforts based on the above?
\end{itemize}

\noindent{}To summarize, our contributions are: 
\begin{enumerate}
    \item Release of PolQA: the first Polish OpenQA dataset,\footnote{PolQA dataset is available at: \url{https://hf.co/datasets/ipipan/polqa}}
    \item Empirical study of data annotation strategies for OpenQA and proposal of an efficient method to create similar datasets for other languages.
    % \item Release of OpenQA models for Polish.\footnote{The OpenQA models are available at: \url{AnonymizedURL}}
\end{enumerate}

% However, it poses a challenge how to efficiently annotate necessary training sets. Historically, the approach was to start with the passage and ask the annotators to come up with a question. This results in biased examples which are easily exploited by the models \cite{}. Moreover, created questions are artificial, while in practice the natural questions are often available (e.g. past customer question, search engine logs, FAQs). In recent works, the annotators are given a set of passages returned by the search engine and are asked to decide which ones are relevant to given question \cite{}. This approach might also lead to the biased training examples as search engines might favor high lexical overlap between question and the passage. In this work, we propose the third approach and give annotators full freedom to find relevant passages.

\section{Related Work}
\label{sec:related-work}
OpenQA is an established task in natural language processing research \citep{zhu2021retrieving}. Over the years, multiple datasets were published and used for training and evaluation of OpenQA systems.

The first version of MS MARCO dataset \citep{nguyen2016ms} consisted of 100,000 questions sampled from Bing's search logs and matching passages obtained from top 10 Bing's search results. Since then the dataset was updated a few times and currently has 1,010,916 questions. Similar to MS MARCO, the contributors of Natural Questions \citep[NQ]{kwiatkowski-etal-2019-natural} sampled 323,045 questions from Google search logs. However, they limited the possible passages to Wikipedia articles. Another OpenQA dataset created using search logs is DuReader \citep{he-etal-2018-dureader}. In contrast to MS MARCO and NQ, DuReader is in Chinese. It makes it the largest (and one of few) OpenQA dataset for the non-English language.

The QA datasets for Polish are more scarce. \emph{Czy wiesz?} dataset  \citep{marcinczuk-etal-2013-evaluation} consists of 4,721 questions from the \emph{Did you know?} section on Polish Wikipedia. However, only 250 of them were manually matched with a relevant passage, the rest was only matched with the whole article. \citet{rybak-etal-2020-klej} extended the aforementioned dataset by labeling passages for additional 1,070 questions. The lack of answers limits the usability of the dataset to training only passage retriever models. The OpenQA system RAFAEL \citep{przybyla2016boosting} used 1,130 questions collected from a Polish quiz TV show \emph{Jeden z dziesięciu} \citep{Karzewski1997}. The dataset contains answers, which allow end-to-end evaluation of the OpenQA system, but lacks question-passage pairs. Two recent resources are related to the PolEval 2021 shared task on QA \citep{ogrodniczukpoleval}. The official dataset contains 6,000 questions together with matching answers. Moreover, one of the participating teams gathered additional 1,000 question-answer pairs \citep{rybak2021retrieve}. Again, both datasets lack matching passages. The latest Polish dataset is PoQuAD \citep{10.1007/978-3-031-21756-2_16,10.1145/3587259.3627548}, a Polish equivalent of SQuAD \citep{rajpurkar-etal-2018-know}. It consists of 70,000 question-answer pairs with passages extracted from Polish Wikipedia. Unlike the PolQA dataset, it was created by asking annotators to write a question about a given passage, rather than by finding the passage to a given question. Often these questions are only valid in the context of the given passage, e.g. "What day did the battle start?", which makes them suitable for reading comprehension but not for neural retrieval.

Few multilingual resources include Polish text. The Cross-lingual OpenQA dataset \citep[XQA]{liu-etal-2019-xqa} consists of an English training set and evaluation data for additional eight languages (including 1,846 questions for Polish). Similar to the \emph{Czy wiesz?} dataset, it was created from the \emph{Did you know?} section of Wikipedia. However, the resulting dataset consists of cloze test statements instead of grammatically correct questions. The Multilingual Knowledge Questions \& Answers \citep[MKQA]{mkqa} contains 10,000 questions sampled from NQ and manually translated into 25 typologically diverse languages. As noted by \citet{rybak2021retrieve}, over 80\% of those questions are not useful for training the OpenQA model as they lack the answer, are ambiguous, or require a long answer (\emph{Why?} and \emph{How?} questions). None of the above resources have matching passages.

\section{Data Collection}
\label{sec:data-collection}
Typical architecture of an open-domain QA system consists of two models -- retriever and reader \citep{zhu2021retrieving}. Retriever finds passages from the corpus that might be relevant to the question. Reader uses those passages to extract (or generate) the final answer. In this section, we describe our approach to annotate triplets of (question, passage, and answer) required to train such systems.

The annotation team consisted of 16 annotators, all native Polish speakers, most of them having linguistic backgrounds and previous experience as annotators. The authors of this study acted as super-annotators, who kept the annotators' work in line with the guidelines. In particular, they reviewed the first 200 labeled examples of each annotator to provide feedback on their work and improve the quality of further annotations. In addition, the super-annotators provided ongoing support during the annotation process, helping with ambiguous examples and clarifying any doubts about the guidelines.

\subsection{Questions and Answers}
\label{sec:qa}
% QA source
The majority of questions come from two existing resources, the 6,000 questions released during the PolEval 2021 shared task on QA \citep{ogrodniczukpoleval} and additional 1,000 questions gathered by one of the shared task participants \citep{rybak2021retrieve}. Originally, the questions come from collections associated with TV shows, both officially published \citep{Karzewski1997} and gathered online by their fans, as well as questions used in actual quiz competitions, on TV or online.

Answers are formulated in a natural language, in a way a Polish speaker would answer the questions. It means that the answers might contain prepositions, be inflected, and contain punctuation. In some cases, the answer might have multiple correct variants, e.g. numbers are written as numerals and words, synonyms, abbreviations and their expansions. We include all such variants.

During the annotation, we cleaned the existing dataset by correcting the factual correctness of questions, adding missing answer variants, and replacing near-duplicates with new questions.

\subsection{Taxonomy}
\label{sec:taxonomy}
We manually classify each question-answer pair based on its (1) formulation, (2) question type, and (3) entity type, according to the taxonomy proposed by \citet{ogrodniczukpoleval}.

\paragraph{Formulation} denotes the kind of expression used to request information. Three types of phrasing are distinguished:\footnote{The examples are translated into English for the convenience of the reader.}
\begin{itemize}
    \item \textbf{plain question}, e.g. \textit{What is the name of the first letter of the Greek alphabet?}
    \item \textbf{command}, e.g. \textit{Expand the abbrev. ’CIA’.}
    \item \textbf{compound}, e.g. \textit{This French writer, born in the 19th century, is considered a pioneer of the sci-fi literature. What is his name?}
\end{itemize}

\paragraph{Question type} indicates what type of information is sought by the question:
\begin{itemize}
    \item \textbf{single entity}, e.g. \textit{Who is the hero in the Tomb Raider video game series?},
    \item \textbf{multiple entities}, e.g. \textit{Which two seas are linked by the Corinth
Canal?},
    \item \textbf{entity choice}, e.g. \textit{Is 'Sombrero' a type of a dance, a hat or a dish?},
    \item \textbf{yes/no}, e.g. \textit{When the term of office of the Polish Sejm is terminated, does it apply to the Senate as well?},
    \item \textbf{other entity name}, e.g. \textit{What was the nickname of Louis I, the King of the Franks?},
    \item \textbf{gap filling}, e.g. \textit{Finish the proverb: 'if you fly with the crows\ldots'}.
\end{itemize}

The question regarding entities can either seek a \textbf{named entity} or an \textbf{unnamed one}. In the former case, the questions are categorised according to the fine-grained \textbf{named entity type}.\footnote{We include 34 types: day, year, century, period, count, quantity, person, name, surname, dynasty, organisation, company, band, country, state, city, nationality, mountain, lake, island, sea, river, range, archipelago, continent, place, vehicle, title, symbol, event, celestial body, animal, building, and other.}

\subsection{Source of Passages}
\label{sec:source-passages}
% passage source
% retrieval
% zero-shot
% hard-negatives
We chose Wikipedia as our source of passages as it contains relevant passages for over 93\% of all questions. The missing questions concern basic arithmetic, proverbs, the content of books or movies, are yes/no questions with a negative answer, or comparison questions, which require multiple passages to answer.

\begin{table*}[!ht]
\renewcommand*{\arraystretch}{1.3}
\setlength{\tabcolsep}{11pt}
\centering
\begin{tabular}{l|rrr|rrr}
    \toprule
    \multirow{2}{*}{\bf{Strategy}} & \multicolumn{3}{c|}{\bf{\# Questions}} & \multicolumn{3}{c}{\bf{\# Passages}} \\
    & \bf{Positive} & \bf{Negative} & \bf{Total} & \bf{Positive} & \bf{Negative} & \bf{Total} \\
    \midrule
    \bf{Standard} & 6,427 & 6,886 & 7,000 & 29,841 & 28,991 & 58,832 \\
    \;\; Manual retrieval & 6,296 & 1,763 & 7,000 & 21,451 & 2,729 & 24,180 \\
    \;\; Neural retrieval & 4,456 & 6,868 & 7,000 &  8,714 & 26,286 & 35,000 \\
    \bf{Efficient} & 5,402 & 6,505 & 7,000 & 13,191 & 21,809 & 35,000 \\
    \midrule
    \bf{Total} & 6,516 & 6,946 & 7,000 & 38,908 & 48,617 & 87,525 \\
    \bottomrule
\end{tabular}
\caption{Number of questions and evidence passages in the PolQA dataset for each annotation strategy. The passage is positive (negative) if it contains (does not contain) an evidence supporting answer. For questions, it refers to the number of questions with at least one positive/negative passage. The number of passages does not sum up because some passages were present in more than one annotation strategy.}
\label{tab:basic-stats}
\end{table*}

\subsection{Candidate Passages}
\label{sec:candidate-passages}
Each Wikipedia article is parsed using WikiExtractor \citep{Wikiextractor2015}, but contrary to default settings we keep lists as a valid text. We split parsed articles into passages at the ends of the paragraphs or if the passage is longer than 500 characters. We try to split on sentence boundaries, whenever possible. Overall, we obtain a knowledge source of 7,097,322 passages.

\subsection{Evidence Passages}
\label{sec:evidence-passages}
We use binary relevance score to annotate passages in the context of asked question. We define a relevant (also called \emph{positive}) passage as a continuous span sentences which allows to answer a question assuming basic reasoning skills (e.g. the conversion of years to centuries) and knowledge (e.g. Poland is a country). Otherwise, the passages is considered irrelevant (or \emph{negative}).

The process of selecting negative passages is crucial for the final performance of neural retrievers. Usually, the best results are achieved when the negative passages are very similar to the given question, i.e. it is hard to decide whether the passage is positive or negative. Therefore, such negatives are often called \emph{hard} negatives. Although there are many methods to select them automatically \citep{ren-etal-2021-rocketqav2,karpukhin-etal-2020-dense}, we decided to evaluate if it is beneficial to label them manually.

\subsection{Annotation Strategies}
\label{sec:strategies}
We follow two strategies to annotate passages for each question and named them \emph{standard} and \emph{efficient}. Each of them consists of two phases: the retrieval of candidate passages and the manual verification of their relevance. In particular, the verification phase is the source of hard negative passages, since passages considered positive in the retrieval phase can be labeled negative in the verification phase.

\paragraph{Standard strategy} The first strategy is to ask annotators to use internal (i.e. Wikipedia Search) or external (e.g. Google) search engines to find a relevant passage in the knowledge source using any keywords or queries they consider useful.

We hypothesize that the unconstrained way of finding the passages will result in more unbiased and diverse examples. Moreover, we ask the annotators to find not one, but up to five passages, preferably from different articles to even further increase passage diversity.

To assure the correctness of the found passages, during the verification phase we double-check each of them to decide if they are relevant (i.e. allow to answer the question) or not. This is the first source of hard negative passages for this strategy. However, since annotators were tasked with finding only relevant passages the number of negatives is relatively small (around 11\% of all labeled passages, see Table \ref{tab:basic-stats}).

To overcome the scarcity of negative passages, we train a neural retriever on the aforementioned passages (see row 3 from Table \ref{tab:results}) and use it to retrieve additional 5 most relevant passages for each question. Then, the annotators manually verified each passage to decide if it is relevant or not. We use irrelevant passages as the second source of hard negatives. 

\paragraph{Efficient strategy}
An alternative approach to annotating passages is to show annotators question-passage pairs and ask them to verify if the passage is relevant or not. This method is several times faster (see Table \ref{tab:annotation}), but it requires a sampling function that selects passages to annotate. Choosing the wrong function can lead to inefficiency (if it selects irrelevant passages) and might bias the dataset (e.g. by selecting passages with high lexical overlap).

We propose the following pipeline as a sampling function. First, we use SpaCy \citep{spacy} to lemmatize questions and passages. Then, the BM-25 algorithm \citep{Robertson2009ThePR} selects top 100 candidate passages which we re-rank using multilingual cross-encoder \citep{bonifacio2021mmarco}.\footnote{\url{https://hf.co/unicamp-dl/mMiniLM-L6-v2-mmarco-v2}} Finally, we select top 5 passages per questions for manual verification.

We purposefully avoid any Polish-specific resources to assure the generality of our approach. Only the lemmatiser is trained on Polish texts, however, it is not required for the method to work and lemmatisers are widely available for many languages \citep{qi-etal-2020-stanza}.

\section{PolQA Dataset}
\label{sec:polqa-dataset}
% statistics of the final dataset
% + counts of questions, passages (positives + negatives) divided based on source
% + split of question based on taxonomy
% + avg len of question, passage, answer
% + token overlap between question and passage
% + how often answer is not directly present
% + annotation time comparison
% + quality comparison
% number of passages per question

\subsection{Data Statistics}
\label{sec:data-stats}

The final PolQA dataset consists of 7,000 questions, 8,713 answer variants, 87,525 evidence passages, and a corpus of 7,097,322 unlabeled candidate passages. For each question, we used two different annotation strategies (see Section \ref{sec:evidence-passages}) to obtain a diverse set of evidence passages. Each strategy results in a different ratio of positive and negative passages (see Table \ref{tab:basic-stats}). By design, the \emph{Standard\textsubscript{manual}} strategy has the highest share of positive passages. Any negatives come from the quality assessment phase during which 11\% of passages were considered not to answer the question (see Table \ref{tab:annotation}). Both \emph{Standard\textsubscript{neural}} and \emph{Efficient} strategy have much more negatives than positives. Overall, 93\% of questions have at least one positive passage and 99\% have at least one negative passage.

\subsection{Question Types}
\label{sec:question-types}
Each question is classified according to three different dimensions (see Section \ref{sec:taxonomy}). Plain questions account for 97.7\% of cases (see Table \ref{tab:taxonomy}). There are 1.5\% of compounds, usually, a statement with an introduction followed by a question. The rest 0.8\% of the questions are commands.

Most questions ask for a single entity (79.8\%) or let to choose a single entity among a few provided (10.1\%). There are 7.6\% yes/no questions, and a small share of other question types.

There is greater variety in entity type, the 55.1\% answers being named entities and 37.3\% of unnamed entities. 
% Full analysis for entity subtypes is presented in Appendix in Table [\ref{tab:taxonomy-subtype}].

\begin{table}
\renewcommand*{\arraystretch}{1.3}
\setlength{\tabcolsep}{6pt}
\centering
\begin{tabular}{l|rr}
    \toprule
    \bf{Category} & \bf{Questions} & \bf{Share} \\
    \midrule
    \multicolumn{3}{c}{\bf{Question formulation}} \\
    \midrule
    Plain question & 6,839 & 97.7\% \\
    Compound &   108 &  1.5\% \\
    Command  &    53 &  0.8\% \\
    
    \midrule
    \multicolumn{3}{c}{\bf{Question type}} \\
    \midrule
    Single entity     & 5,589 & 79.8\% \\
    Entity choice     &   705 & 10.1\% \\
    Yes/No            &   532 & 7.6\% \\
    Other name        &    86 & 1.2\% \\
    Multiple entities &    59 & 0.8\% \\
    Gap filling       &    29 & 0.4\% \\
    
    \midrule
    \multicolumn{3}{c}{\bf{Entity type}} \\
    \midrule
    Named   & 3,854 & 55.1\% \\
    Unnamed & 2,614 & 37.3\% \\
    Yes/No  &   532 &  7.6\% \\

    \bottomrule
\end{tabular}
\caption{Distribution of questions based on their formulation, type, or type of entity.}
\label{tab:taxonomy}
\end{table}

\subsection{Lexical Similarity}
\label{sec:lexical-similarity}
One of the main limitations of the traditional retrieval algorithms (e.g. BM-25) is their dependence on the lexical overlap. If there is no common token between the question and the relevant passage then it won't be found by the retriever. However, the OpenQA datasets are often created by automatically finding the candidate passages first and then asking annotators to assess if they are positive or not (see Section \ref{sec:related-work}). This might introduce bias in the final dataset as the method of finding candidate passages might rely on the lexical overlap. As a result, the lexical methods will perform well on such a dataset.

We analyze the similarity between questions and passages by calculating both token (before lemmatisation) and lemma overlap as a percentage of question tokens/lemmas (excluding punctuation) that appear in the matching passage (see Table \ref{tab:overlap}). As expected, the \emph{Standard\textsubscript{manual}} strategy of finding passages has a much lower lemma overlap for positive passages (35.6\%) than \emph{Efficient} method which used a lexical retriever to find candidate passages (51.5\%). Interestingly, even the negative passages obtained with \emph{Efficient} method have higher overlap (47.9\%) which makes them perfect as hard negatives for training a neural retriever. The \emph{Standard\textsubscript{neural}} strategy sits in between those two methods, the usage of neural retriever leads to higher overlap for positive methods (42.8\%) and comparable for negative ones (27.8\%).

The same type of analysis is beneficial for comparing passages and the answers. We calculate both token and lemma overlap to understand how often the answer is directly present in the passage (see Table \ref{tab:overlap}) and can be simply extracted by the reader and when it has to be generated. We excluded the yes/no questions since in those cases any overlap would be accidental and meaningless. The annotation strategy has a minimal influence on the overlap for positive passages with \emph{Efficient} method having the highest overlap and \emph{Standard\textsubscript{neural}} the lowest. For \emph{Standard\textsubscript{manual}} strategy, the overlap between positive and negative passages is similar since annotators were asked to only find positive passages. For other methods, the negative passages have much lower overlap, i.e. do not contain answers. The token overlap of around 40\% indicates the high difficulty of the PolQA dataset. The reader cannot fall back on copying the correct span of the passage but has to generate an answer containing novel tokens. Partially, the difficulty is solved by learning how to lemmatise but even the lemma overlap is still only around 70\%.

\begin{table}
\renewcommand*{\arraystretch}{1.3}
\setlength{\tabcolsep}{5.5pt}
\centering
\begin{tabular}{l|cc|cc}
    \toprule
    \multirow{2}{*}{\bf{Strategy}} & \multicolumn{2}{c}{\bf{Token overlap}} & \multicolumn{2}{c}{\bf{Lemma overlap}} \\
    & \bf{Pos} & \bf{Neg} & \bf{Pos} & \bf{Neg} \\
    \midrule
    \multicolumn{5}{c}{\bf{Questions}} \\
    \midrule
    \bf{Standard} & & \\
    \;\; Manual & 15.5\% & 12.5\% & 35.6\% & 28.7\% \\
    \;\; Neural & 20.3\% & 13.1\% & 42.8\% & 27.8\% \\
    \bf{Efficient} & 25.1\% & 23.1\% & 51.5\% & 47.9\% \\
    \midrule
    \multicolumn{5}{c}{\bf{Answers}} \\
    \midrule
    \bf{Standard} & & \\
    \;\; Manual & 41.1\% & 33.8\% & 71.5\% & 60.8\% \\
    \;\; Neural & 40.1\% & 10.4\% & 70.2\% & 16.3\% \\
    \bf{Efficient} & 45.0\% & 14.5\% & 75.4\% & 22.9\% \\
    \bottomrule
\end{tabular}
\caption{Average token/lemma overlap between questions/answers and positive (\emph{Pos}) or negative (\emph{Neg}) passages. We calculate token/lemma overlap as a percentage of question/answer tokens/lemmas (excluding punctuation) that appear in the matching passage. We take a maximum if there is more than one answer. We exclude questions with \emph{yes/no} answers which might have only accidental overlap.}
\label{tab:overlap}
\end{table}

\subsection{Annotation Quality}
\label{sec:annotation-performance}

There is a significant time difference between freely searching for a passage (\emph{Standard\textsubscript{manual}} method) and verifying if the passage is correct (\emph{Efficient} and \emph{Standard\textsubscript{neural}} methods). On average, searching for the passage takes 75.5 seconds per passage while verifying one is over five times faster and takes only 13.6 seconds (see Table \ref{tab:annotation}). There is no significant difference between the correctness of passages obtained by both methods with a precision of around 89\% and a recall of around 96\%.

\begin{table}
\renewcommand*{\arraystretch}{1.3}
\setlength{\tabcolsep}{6pt}
\centering
\begin{tabular}{l|rrr}
    \toprule
    \bf{Task} & \bf{Time} & \bf{Precision} & \bf{Recall} \\
    \midrule
    Free search & 75.5 & 88.7\% & 96.6\% \\
    Verification & 13.6 & 89.9\% & 95.8\% \\
    \bottomrule
\end{tabular}
\caption{Data annotation statistics depending on the type of the annotation task, freely searching for a passage or verifying if a given passage is correct. Time refers to the average time in seconds to annotate one question-passage pair.}
\label{tab:annotation}
\end{table}

\begin{table*}[!ht]
\renewcommand*{\arraystretch}{1.3}
\setlength{\tabcolsep}{6pt}
\centering
\begin{tabular}{ll|ccc|ccc}
    \toprule
    \bf{\#} & \bf{Strategy} & \bf{Verified} & \bf{\# Positives} & \bf{\# Hard Negatives} & \bf{Retriever} & \bf{Reader} & \bf{E2E} \\
    \midrule
    1 & Standard & No & Single & No & 52.57\% & 75.33\% & 46.54\% \\
    2 & Standard & Yes & Single & No & 52.90\% & 72.99\% & 44.42\% \\
    3 & Standard & Yes & All & No & 53.24\% & 80.25\% & 47.77\% \\
    4 & Standard & Yes & All & Yes & 46.76\% & \bf{80.92\%} & 51.23\% \\
    \midrule
    5 & Efficient & No & All & No & 54.24\% & 77.68\% & 51.23\% \\
    6 & Efficient & Yes & All & No & 59.26\% & 79.46\% & 52.46\% \\
    7 & Efficient & Yes & All & Yes & \bf{61.16\%} & 77.79\% & \bf{56.25\%} \\
    \bottomrule
\end{tabular}
\caption{OpenQA model performance trained on passages obtained using different annotation strategies. We use the top 10 accuracy on the validation set. Verified refers to whether the passage was additionally verified by the annotator or taken as-is. Positives refers to the number of passages per question. Hard Negatives refers to whether hard negatives (passages manually verified as negatives) were used in training.}
\label{tab:results}
\end{table*}

\section{Evaluation}
\label{sec:evaluation}
We use standard retriever-reader architecture to train an OpenQA system and evaluate different annotation strategies. We split the dataset into the train (5,000 questions, 27,131 relevant passages), validation (1,000 questions, 5,839 relevant passages), and test (1,000 questions, 5,938 relevant passages) sets. To avoid potential bias introduced by the annotation approach, we limit the validation and test sets to passages found using \emph{Standard\textsubscript{manual}} method (3,160 and 3,252 relevant passages respectively). The experiments use the validation set to evaluate the models, except for the final evaluation (Section \ref{sec:summary}), where the test set is employed. We use a single V100 GPU for all experiments.

\subsection{Retriever}
\label{sec:retriever}
As a retriever, we use HerBERT Base model \citep{mroczkowski-etal-2021-herbert} and fine-tune it with a triplet loss \citep{weinberger2009distance} and a margin of 0.1. We train this encoder with AdamW optimizer \citep{Loshchilov2019DecoupledWD} for 50 epochs using a learning rate of $10^{-5}$ and a linear decay schedule. During the training, we sample one positive and one negative passage for each question. We randomly sample negatives from a training set, except for the experiments in Section \ref{sec:hard-negatives}, where manually labeled negative passages are used instead. In all cases, we also use in-batch negatives.

During the evaluation, we first encode all Wikipedia passages and index them using FAISS \citep{johnson2019billion}. Then, for each question, we retrieve the top 10 most similar passages through an exhaustive search. We measure model performance through the accuracy of top 10 candidates.

\subsection{Reader}
\label{sec:reader}
As a reader, we use plT5 Base model \citep{chrabrowa2022evaluation} and train it in a text-to-text mode. We concatenate the question with all available relevant\footnote{Except for the experiments in Section \ref{sec:hard-negatives}, where manually labeled negative passages are also used.} passages as input and generate a sequence of tokens as an answer. We fine-tune the model with AdaFactor optimizer \citep{shazeer2018adafactor} for 10 epochs using a learning rate of $10^{-4}$ and a linear decay schedule.

To evaluate the reader (as well as the end-to-end system), we use the metric proposed by \citet{ogrodniczukpoleval}. For numerical answers, we extract the numeral (Arabic or Roman) using regular expression and expect the equality between prediction and true value. For the rest of the questions, we calculate character-wise Levenshtein distance \citep{1965}, which is allowed to reach 50\% of the answer length for a match. In case there is more than one correct answer, we compare the prediction to each and choose the best matching ones.

We evaluate the reader both on manually labeled positive passages (to assess the reader quality itself), as well as on passages returned by the retriever for an end-to-end evaluation.

\section{Results}
\label{sec:results}
To understand how different annotation strategies influence the performance of an OpenQA system, we conduct a series of experiments to answer the aforementioned research questions (see Section \ref{sec:introduction}):
\begin{itemize}
    \item In Section \ref{sec:quality-assessment} and \ref{sec:human-vs-zero}, we investigate the benefit of high-quality and unbiased training data (RQ1).
    \item In Section \ref{sec:number-of-passages}, we analyze how the performance improves with the increased number of training passages (RQ2).
    \item In Section \ref{sec:hard-negatives}, we compare models with and without the annotated hard negative passages (RQ3).
    \item In Section \ref{sec:summary}, we summarize our experiments and compare two annotation strategies based on their cost (in terms of human effort) and impact on OpenQA performance (RQ4, RQ5).
\end{itemize}

\subsection{Quality Assessment}
\label{sec:quality-assessment}
% [v1] task A, single passage
% [v2] task A + verification from task B, single passage

We train a model using all passages from \emph{Standard\textsubscript{manual}} strategy and compare it to the model using only verified ones. Even though the precision of human passages is around 89\% (see Table \ref{tab:annotation}) the impact of additional verification on retriever performance is negligible (see rows 1, 2 in Table \ref{tab:results}). Using only verified passages actually decreases the accuracy of the reader by 2.34 p.p. It has a similar impact on the end-to-end score (a decrease from 46.54\% to 44.42\%).

\subsection{Number of Relevant Passages}
\label{sec:number-of-passages}
% [v2] task A + verification from task B, single passage
% [v4] task A + verification from task B

The next design choice in the \emph{Standard\textsubscript{manual}} strategy is how many relevant passages should be found for each question. Increasing this number heavily influences the annotators' workload, since every next passage for a given question usually takes more time to find. We compare models trained with a single or all (i.e. up to five) passages per question (see rows 2, 3 in Table \ref{tab:results}). In both cases, we use only verified passages. Again, the difference for the retriever is positive but small (52.9\% vs 53.24\%). For the reader, the number of used passages is more important. The model trained on all passages scores 7.25 p.p. higher compared to one trained on only one passage per question. The impact of the number of passages on the end-to-end accuracy is also positive as it increases the performance by 3.35 p.p.

\subsection{Retrieval vs Verification}
\label{sec:human-vs-zero}
% [v4] task A + verification from task B
% [b3] zero-shot (no training) (?)
% [v5] zero-shot (training)
% [v6] zero-shot from task B (without negatives)

% Q: should we filter zero-shot with reader
% A: probably not since we don't have data to train it
We compare two main strategies to annotate the passages, freely searching for relevant passages (\emph{Standard\textsubscript{manual}}) or manual verification of passage candidates (\emph{Efficient}). Additionally, we train a model on passages from \emph{Efficient} but without any manual annotation, i.e. treating all passages returned by the sampling function as positives. This can be viewed as a simple model distillation \citep{ren-etal-2021-rocketqav2}.

Surprisingly, the retriever trained on unlabeled examples obtained using \emph{Efficient} strategy performs better than the retriever trained on manually annotated data (see rows 3, 5 in Table \ref{tab:results}). It shows that thanks to the high generalizability of cross-encoders it is possible to use a multilingual model to automatically find relevant passages and use them to create a high-quality dataset.

If we additionally verify examples obtained with \emph{Efficient} strategy, we get an additional 5 p.p. improvement in retriever performance (see rows 5, 6 in Table \ref{tab:results}). The manual annotation increases also the accuracy of the reader (from 77.68\% to 79.46\%) and the whole system (from 51.23\% to 52.46\%). However, the best reader performance (80.25\%) is achieved by training on \emph{Standard\textsubscript{manual}} passages.

\subsection{Hard Negatives}
\label{sec:hard-negatives}
% [v4] task A + verification from task B
% [v7] task A + verification and hard-negatives from task B
% [v6] zero-shot from task B (without negatives)
% [v8] zero-shot with negatives

We explore the impact of hard negatives on model performance by using manually annotated negative passages instead of randomly sampled ones. For \emph{Standard} strategy, we additionally use all passages obtained with \emph{Standard\textsubscript{neural}} method (see Section \ref{sec:evidence-passages}). For \emph{Efficient} we simply include the passages annotated as negatives.

Including hard negatives in \emph{Standard} dataset decreases the retriever performance from 53.24\% to 46.76\% (see rows 3, 4 in Table \ref{tab:results}). The opposite happens for the reader. Additional passages slightly improve the accuracy of the reader (by 0.67 p.p.) which leads to an end-to-end increase of 3.46 p.p.

For the \emph{Efficient} method, the hard negatives improve the retriever performance by almost 2 p.p. and at the same time hurt the reader performance by almost 2 p.p. However, the end-to-end accuracy increases from 52.46\% to 56.25\% (see rows 6, 7 in Table \ref{tab:results}).

%TODO comment

\subsection{Summary}
\label{sec:summary}
% [v7] "standard" annotation
% [v8] "efficient" annotation
% [v9] all data
To summarize, we compare two data annotation strategies. In \emph{Standard} approach, we use all manually verified \emph{Standard\textsubscript{manual}} passages for training retriever. For reader, we additionally include all \emph{Standard\textsubscript{neural}} passages. The \emph{Efficient} approach uses verified passages (both positives and negatives) from \emph{Efficient} method.

\begin{table}
\renewcommand*{\arraystretch}{1.3}
\setlength{\tabcolsep}{4pt}
\centering
\begin{tabular}{l|rrrr}
    \toprule
    \bf{Strategy} & \bf{Retriever} & \bf{Reader} & \bf{E2E} & \bf{Time} \\
    \midrule
    Standard  & 51.47\% & 78.45\% & 51.47\% & 376 \\
    Efficient & 62.02\% & 74.43\% & 53.21\% & 68 \\
    % \midrule
    % Combined & 62.02\% & 78.45\% & 53.10\% & 444 \\
    \bottomrule
\end{tabular}
\caption{OpenQA model performance trained on passages obtained using standard and efficient annotation strategy. We use the top 10 accuracy on the test set. Time refers to the average time in seconds to annotate passages for one question.}
\label{tab:summary}
\end{table}

For retrievers, the model trained on passages obtained with \emph{Efficient} strategy results in 10 p.p. higher accuracy compared to \emph{Standard} approach (see Table \ref{tab:summary}). For readers, the \emph{Standard} approach works better by almost 4 p.p. The end-to-end for both methods are similar, the \emph{Efficient} method has an accuracy of 53.21\% compared to 51.47\% for \emph{Standard} annotation strategy. Although the final results are similar, the time spent annotating the data is very different. The \emph{Efficient} approach requires over five times less time to annotate passages for a single question (68 vs 376 seconds).

% These results suggest that if we combine retriever trained on passages obtained with \emph{Efficient} method and reader trained on \emph{Standard} passages, we might end up with the best end-to-end results. However, this is not the case and the accuracy is lower than for the model trained using only \emph{Efficient} passages.

\section{Conclusion}
\label{sec:conclusion}

In this work, we present PolQA, the first Polish dataset for OpenQA. It consists of 7,000 questions together with 8,713 answer variants and 87,525 evidence passages obtained by different methods to increase their diversity and completeness.

This resource allows us to evaluate the performance of the OpenQA model depending on different data annotation strategies and formulate the following recommendations for creating a similar OpenQA dataset for other languages:
\begin{itemize}
    \item Obtaining unbiased evidence passages does not improve the performance of OpenQA models. Instead, we recommend using \emph{Efficient} strategy to sample candidate passages and manually verify their correctness. This reduces the cost of annotation over five times and at the same time increases the performance of the OpenQA model (see Section \ref{sec:human-vs-zero}).
    \item If the annotation cost is still a limiting factor, then using unlabeled passages retrieved by the sampling function from \emph{Efficient} strategy is a competitive (or even better) strategy than obtaining unbiased passages but requires no manual annotation (see Section \ref{sec:human-vs-zero}).
    \item It is beneficial to annotate multiple evidence passages per question, as well as to include not only positive but also negative passages (see Section \ref{sec:number-of-passages} and \ref{sec:hard-negatives}).
    \item Depending on the experience and skill of the annotators, it might not be necessary to double-check their work (see Section \ref{sec:quality-assessment}).
\end{itemize}

We hope our work will enable research on Polish OpenQA and be beneficial to the wider OpenQA research community, both to researchers working on cross-lingual OpenQA and those who seek an efficient approach to create OpenQA datasets.

\section{Acknowledgments}
This work was supported by the European Regional Development Fund as a part of 2014–2020 Smart Growth Operational Programme, CLARIN — Common Language Resources and Technology Infrastructure, project no. POIR.04.02.00-00C002/19.

\nocite{*}
\section{Bibliographical References}\label{reference}
\bibliographystyle{lrec_natbib}
\bibliography{lrec-coling2024-example}

% \section{Language Resource References}
% \bibliographystylelanguageresource{lrec_natbib}
% \bibliographylanguageresource{languageresource}

\end{document}